\documentclass[runningheads]{llncs}
\usepackage{graphicx} 
\usepackage[T1]{fontenc}
\usepackage[utf8x]{inputenc}
\usepackage[english]{babel}
\usepackage[title]{appendix}

\usepackage{amsmath}
\usepackage{amssymb}
\usepackage{mathtools}

\makeatletter
\newcommand{\@chapapp}{\relax}%
\makeatother

\def\keywordname{{\bf Keywords:}}%

\title{Vulnerability Detection via Topological Analysis of Attention Maps}
\author{Snopov P. \inst{1},
Golubinsky A.N.\inst{1}}

\institute{Institute for Information Transmission Problems of Russian Academy of Sciences
\email{snopov@iitp.ru}\\
}

\begin{document}

\maketitle
\begin{abstract}
    Recently, deep learning (DL) approaches to vulnerability detection have gained significant traction. These methods demonstrate promising results, often surpassing traditional static code analysis tools in effectiveness.

    In this study, we explore a novel approach to vulnerability detection utilizing the tools from topological data analysis (TDA) on the attention matrices of the BERT model. Our findings reveal that traditional machine learning (ML) techniques, when trained on the topological features extracted from these attention matrices, can perform competitively with pre-trained language models (LLMs) such as CodeBERTa. This suggests that TDA tools, including persistent homology, are capable of effectively capturing semantic information critical for identifying vulnerabilities.

    \end{abstract}

\section{Introduction}
The problem of source code vulnerability detection has become increasingly 
important in the field of software engineering. This complex task involves analyzing 
both the syntax and semantics of the source code. Traditionally, static analysis 
methods have dominated the field of vulnerability detection. 
These static tools tend to rely heavily on the syntax of the program, thereby 
overlooking potential information hidden in the semantics. As a result, 
such approaches often suffer from high rates of false positives.

Recently, however, various machine learning (ML) and deep learning (DL) methods have 
emerged \cite{sharma_survey_2024,steenhoek_empirical_2023}. Most DL-based solutions 
leverage graph neural networks (GNNs) and large language models (LLMs). 
These methods aim to learn both the semantics and syntax of the program. 
They have shown promising results, outperforming traditional static analysis tools and 
demonstrating low rates of false positives and false 
negatives \cite{katsadouros_survey_2022}.

Our work presents a novel approach to the vulnerability detection problem.
As previously mentioned, the superiority of DL-based models over static ones is 
potentially due to their ability to capture the semantic information of the program. 
This semantic information is captured by the neural networks during training or 
fine-tuning. Furthermore, LLM-based models can capture this information even without 
fine-tuning.

On the other hand, the semantics of the source code, like its syntax, can also be represented as a 
tree or a graph. This representation allows for the application of topological methods.

In this work, we leverage the BERT model, pre-trained on source code, to capture the 
semantic information of programs. We also utilize tools from topological 
data analysis (TDA) \cite{tda4ds} to explore the relationships within this semantic 
information. Our aim is to analyze and interpret the attention matrices using 
topological instruments.

Our work is inspired by the work of Laida Kushareva et. al. \cite{kushnareva_artificial_2021}, 
in which the authors apply the topological methods for the artificial text detection. They demonstrated that 
the topology of the semantics captured by an LLM model (such as BERT) provides 
sufficient information for successful classification. Their approach outperformed 
neural baselines and performed on par with a fully fine-tuned BERT model, while being 
more robust to unseen data.

\section{Background}
\subsection{BERT Model}
BERT (Bidirectional Encoder Representations from Transformers) is a 
transformer-based language model that has set new benchmarks in various natural 
language processing (NLP) tasks. The BERT architecture
consists of $L$ encoder layers, each with $H$ attention heads.
Each attention head receives a matrix $X$ as input, representing the 
$d$-dimensional embeddings of $m$ tokens, so that 
$X$ is of shape $m\times d$. The head outputs an updated 
representation matrix $X^{\textrm{out}}$: 
\begin{equation}
    X^\mathrm{out} = W^\mathrm{attn}(XW^V), \textrm{ with } 
    W^\mathrm{attn} = \mathrm{softmax}\bigl( \frac{(XW^Q)(XW^K)^T}{\sqrt{d}} \bigr).
\end{equation}
Here $W^Q, W^K, W^V$ are trained projection matrices of shape $d\times d$
and $W^\mathrm{attn}$ is matrix of attention weights with shape $m\times m$ 
Each element $w_{ij}^\mathrm{attn}$ can be interpreted as a weight of the
$j$-th input's {\it relation} to the $i$-th output where larger weights 
indicate a stronger connection between the two tokens.

\subsection{Attention Graph}
An {\it attention graph} is a weighted graph representation of an attention matrix $W^{attn}$, 
where vertices represent the tokens and the edges connect a pair of tokens if the 
corresponding weight exceeds a predefined threshold value.

Setting this threshold value is critical yet challenging, as it distinguishes weak 
and strong relations between tokens. Additionally, varying the threshold can 
significantly alter the graph structure.

Topological data analysis (TDA) methods can extract properties of the graph structure 
without specifying an exact threshold value, addressing this challenge effectively.

\subsection{Topological Data Analysis}
Topological data analysis (TDA) is an emerging field that applies 
algebraic topology methods to data science. There are numerous excellent tutorials 
and surveys available for both 
non-mathematicians \cite{murugan2019introductiontopologicaldataanalysis,tda4ds} and 
those with a mathematical background \cite{Edelsbrunner2008,Oudot2015,Schenck2022}.

The main tool in topological data analysis, {\it persistent homology}, 
tracks changes in the topological structure of various objects, such as point clouds, 
scalar functions, images, and weighted graphs \cite{adams2016persistenceimagesstablevector,Aktas2019}.

In our work, given a set of tokens $V$ and an attention matrix $W$,we construct a 
family of attention graphs indexed by increasing threshold values. This family, known 
as a {\it filtration}, is a fundamental object in TDA

With this sequence of graphs, we compute persistent homology in dimensions $0$ and $1$.
Dimension $0$ reveals connected components or clusters in the data, while 
dimension $1$ identifies cycles or <<loops>>. 
These computations yield a {\it persistence diagram}, which can be used to derive 
specific topological features, such as the number of connected components and 
cycles (such features are also called the {\it Betti numbers}) 
(see Appendix \ref*{persistent_homology} for the details).

\section{Topological Features of the Attention Graphs}
\label{features}
For each code sample, we calculate the persistent homology in dimensions $0$ and $1$ of 
the symmetrized attention matrices, obtaining the persistence diagram for each 
attention head of the BERT model. We compute the following features in each dimension 
from the diagrams:
\begin{itemize}
    \item Mean lifespan of points on the diagram
    \item Variance of the lifespan of points on the diagram
    \item Maximum lifespan of points on the diagram
    \item Total number of points on the diagram
    \item Persistence entropy
\end{itemize}

We symmetrize attention matrices to enable the application of persistent homology techniques. 
Symmetrizing attention matrices allows us to interpret them as distance matrices of a point cloud embedded in Euclidean space. 
We symmetrize attention matrices as follows: 
\begin{equation}
    \forall i,j: W^\mathrm{sym}_{ij} = \max{(W_{ij}^\mathrm{attn}, W_{ji}^\mathrm{attn})}.
\end{equation}

Alternatively, one can think of attention graphs, in which an edge between the vertices $i$ and $j$
appears if the threshold is greater than both $W_{ij}^\mathrm{attn}$ and $W_{ji}^\mathrm{attn}$.

We consider these features asthe numerical characteristics of the semantic evolution processes 
in the attention heads. These features encode the information about the clusters of mutual influence
of the tokens in the sentence and the local structures like cycles. The features with <<significant>> 
persistence (i.e. those with large lifespan) correspond to the stable processes, whileas the features
with short lifespans are highly susceptible to noise and do not reflect the stable topological attributes.

\section{Experiments}
\subsubsection*{Methodology} 
To evaluate whether the encoded topological information can be used for 
vulnerability detection, we train Logistic Regression, Support Vector Machine (SVM), 
and Gradient Boosting classifiers on the topological features derived from the 
attention matrices of the BERT model, as described in Section \ref*{features}. 
We utilize the {\it scikit-learn} library \cite{scikit-learn} for Logistic Regression 
and SVM, and the {\it LightGBM} library \cite{ke2017lightgbm} for Gradient Boosting. 
Detailed training procedures are outlined in Appendix \ref*{training_details}.

\subsubsection*{Data} 
We train and evaluate our classifier on {\it Devign} dataset.
This dataset comprises samples from two large, widely-used open-source C-language projects: QEMU, and FFmpeg, 
which are popular among developers and diverse in functionality.
Due to computational constraints, we were only using those data samples, that, 
being tokenized, are of length less than $150$. This ensures that the point cloud 
constructed during attention symmetrization is also limited to a maximum length of $150$.

\subsubsection*{Baselines} 
We employ the \texttt{microsoft/codebert-base} model \cite{feng2020codebert} from the 
HuggingFace library \cite{huggingface} as our pre-trained BERT-based baseline. 
Additionally, we fully fine-tune the \texttt{microsoft/codebert-base} model for comparison.
\section{Results and Discussion}
Table \ref*{results} outlines the results of the vulnerability detection experiments 
on the {\it Devign} dataset. The results reveal that the proposed topology-based 
classifiers outperform the chosen large language model (LLM) without fine-tuning but 
perform worse than the fine-tuned version.

\begin{table}
    \label{tab:results}
    \centering
    \caption{The results of the vulnerability detection experiments.}\label{results}
    \begin{tabular}{|c|c|c|}
    \hline
    {\bf Model} & {\bf F1 score} & {\bf Accuracy} \\
    \hline
    Logistic Regression & 0.22 & 0.54 \\
    \hline
    LightGBM & {\bf 0.55} & 0.63 \\
    \hline
    SVM & 0.54 & {\bf 0.65} \\
    \hline
    CodeBERTa (pre-trained) & 0.28 & 0.45 \\
    \hline
    CodeBERTa (fine-tuned) & {\bf 0.71} & {\bf 0.72} \\
    \hline
    \end{tabular}
\end{table}

These observations indicate that the information about a code snippet's vulnerability 
is encoded in the topological attributes of the attention matrices. The semantic 
evolution in the attention heads reflects code properties that are crucial for the 
vulnerability detection task, and persistent homology proves to be an effective method 
for extracting this information.

Notably, only the semantic information from attention heads was used. 
The inclusion of additional topological features obtained from the structural 
information of the source code, such as the topology of graph representations of the 
source code, could potentially enhance the overall performance of the proposed models.
\section{Consclusion}
This paper introduces a novel approach for the vulnerability detection task based on 
topological data analysis (TDA). We propose a set of interpretable topological 
features, obtained from persistence diagrams derived from the attention matrices of 
any transformer-based language model. The experiments demonstrate that machine 
learning classifiers trained on these features outperform pre-trained 
code-specific large language models (LLMs).

Our code is publicly available\footnote{https://github.com/Snopoff/Vulnerability-Detection-via-Topological-Analysis-of-Attention-Maps}, and we encourage further research into TDA-based methods for vulnerability detection and other NLP tasks. Future work could explore combining topological features that encode semantics with those capturing structural information. Additionally, studying different symmetrizations of attention matrices and how they encode semantics could provide new insights. Another interesting direction is applying multiparameter persistent homology to analyze the semantic evolution in attention heads.
\subsubsection*{Acknowledgements}
The work was carried out as part of the state assignment N 1021061609772-0-1.2.1
(FFNU-2024-0020).

\bibliographystyle{splncs04}
\bibliography{biblio}

\begin{subappendices}
    \newpage
    \renewcommand{\thesection}{\Alph{section}}%
    \section{Training Details}
    \label{training_details}
    \subsubsection*{Fine-tuning CodeBERTa} We trained with the cosine scheduler 
    with an initial learning rate $\mathrm{lr}=5e-5$ and set the number of epochs $\mathrm{e}=15$.
    \subsubsection*{Hyperparameter search} We employed {\it Optuna} \cite{optuna} to 
    find optimal hyperparameters for both SVM and LightGBM models.
    
    For SVM, the optimal hyperparameters were
    $C=9.97$, $\gamma=\mathrm{auto}$ and $\mathrm{kernel}=\mathrm{rbf}$. 
    
    For LightGBM, the optimal parameters were
    $\lambda_{\ell_1}=5.97$, $\lambda_{\ell_2}=0.05$, $\mathrm{num\_leaves}=422$, $\mathrm{feature\_fraction}=0.65$,
    $\mathrm{bagging\_fraction}=0.93$, $\mathrm{bagging\_freq}=15$, $\mathrm{min\_child\_samples}=21$.
    \section{Persistent Homology}
    \label{persistent_homology}
    Recall that a simplicial complex $X$ is a collection of $p$-dimensional simplices, 
    i.e., vertices, edges, triangles, tetrahedrons, and so on. Simplicial complexes 
    generalize graphs, which consist of vertices ($0$-simplices) and edges 
    ($1$-simplices) and can represent higher-order interactions.
A family of increasing simplicial complexes

\[
\emptyset \subseteq X_0 \subseteq X_1 \subseteq \cdots \subseteq X_{n-1} \subseteq X_n
\]

is called a \textit{filtration}.

The idea of \textit{persistence} involves tracking the evolution of simplicial complexes over the filtration. 
{\it Persistent homology} allows to trace the changes in homology vector 
spaces\footnote{Usually, homology groups are considered with integral coefficients, but in the realm of 
persistence, homology groups are taken with coefficients in some field, for example, $\mathbb{Z}_p$ for some large prime number $p$. Hence, the homology groups become homology vector spaces.} of simplicial complexes 
that are present in the filtration. Given a filtration $\{X_t\}_{t=0}^n$, the homology functor $H_p$ applied 
to the filtration generates a sequence of vector spaces $H_p(X_t)$ and maps $i_*$ between them

\[
H_p(X_*): H_p(X_0) \xrightarrow{i_0} H_p(X_1) \xrightarrow{i_1} \cdots \xrightarrow{i_{n-1}} H_p(X_n).
\]

Each vector spaces encodes information about the simplicial complex $X$ and its 
subcomplexes $X_i$. For example, $H_0$ generally encodes the connectivity of the space 
(or, in data terms, $H_0$ encodes the clusters of data), $H_1$ encodes the presence of 1-cycles, i.e., loops, 
$H_2$ represents the presence of 2-cycles, and so on. Persistence tracks the generators of each 
vector space through the induced maps. Some generators will vanish, while others will persist. 
Those that persist are likely the most important, as they represent features that truly exist in $X$. 
Therefore, persistent homology allows one to gain information about the underlying topological space via 
the sequence of its subspaces, the filtration.

In algebraic terms, the sequence of vector spaces $H_p(X_t)$ and maps $i_*$ between them can be seen as a 
representation of a quiver $I_n$. From the representation theory of quivers 
it is known, due to Gabriel, that any such representation is isomorphic to a direct sum of indecomposable 
interval representations $I[b_i, d_i]$, that is,

\[
H_p(X_*) \simeq \bigoplus_i I[b_i, d_i].
\]

The pairs $(b_i, d_i)$ represent the persistence of topological features, where $b_i$ denotes the time of 
birth and $d_i$ denotes the time of death of the feature. These pairs can be visualized via {\it barcodes} 
where each bar starts at $b_i$ and ends at $d_i$.

This information is also commonly represented using {\it persistence diagrams}. A persistence diagram is a 
(multi)set of points in the extended plane $\overline{\mathbb{R}^2}$, which reflects the structure of persistent homology.
 Given a set of pairs $(b_i, d_i)$, each pair can be considered as a point in the diagram with coordinates $(b_i, d_i)$. Thus, a 
 persistence diagram is defined as

\[
\mathrm{dgm}(H_p(X_*)) = \{ (b_i, d_i) : I[b_i, d_i] \text{ is a direct summand in } H_p(X_*) \}.
\]

Persistence diagrams provide a detailed visual representation of the topology of 
point clouds. However, integrating them into machine learning models presents 
significant challenges due to their complex structure. To effectively use the 
information from persistence diagrams in predictive models, it is crucial to 
transform the data into a suitable format, such as by applying persistence entropy.

{\it Persistence entropy} is a  specialized form of Shannon entropy specially designed for persistence
diagrams and is calculated as follows:

\[
PE_k(X) \coloneqq -\sum_{(b_i, d_i) \in D_k} p_i \log(p_i),
\]

where

\[p_i = \frac{d_i - b_i}{\sum_{(b_i, d_i) \in D_k} (d_i - b_i)} \quad \text{and} \quad D_k \coloneqq \mathrm{dgm}(H_k(X_\bullet)).\]

This numerical characteristic of a persistence diagram has several advantageous 
properties. Notably, it is stable under certain mild assumptions. This stability means 
there is a bound that <<controls>> the perturbations caused by noise in the input data.
\end{subappendices}

\end{document}